\documentclass{article}
\usepackage{spconf,amsmath,epsfig}

\usepackage{times}
\usepackage{epsfig}
\usepackage{graphicx}
\usepackage{amsmath}
\usepackage{amssymb}

\usepackage{bm}
\usepackage{subfigure}
\usepackage{graphicx}
\usepackage{caption}

\usepackage{algorithm}
\usepackage{algpseudocode}
\usepackage{url}
\makeatletter
\newcommand*{\algrule}[1][\algorithmicindent]{\makebox[#1][l]{\hspace*{.5em}\thealgruleextra\vrule height \thealgruleheight depth \thealgruledepth}}%
\newcommand*{\thealgruleextra}{}
\newcommand*{\thealgruleheight}{.75\baselineskip}
\newcommand*{\thealgruledepth}{.25\baselineskip}

\newcount\ALG@printindent@tempcnta
\def\ALG@printindent{%
	\ifnum \theALG@nested>0
	\ifx\ALG@text\ALG@x@notext
	\else
	\unskip
	\addvspace{-1pt}
	\ALG@printindent@tempcnta=1
	\loop
	\algrule[\csname ALG@ind@\the\ALG@printindent@tempcnta\endcsname]%
	\advance \ALG@printindent@tempcnta 1
	\ifnum \ALG@printindent@tempcnta<\numexpr\theALG@nested+1\relax
	\repeat
	\fi
	\fi
}%
\usepackage{etoolbox}
\patchcmd{\ALG@doentity}{\noindent\hskip\ALG@tlm}{\ALG@printindent}{}{\errmessage{failed to patch}}
\makeatother

\newbox\statebox
\newcommand{\myState}[1]{%
	\setbox\statebox=\vbox{#1}%
	\edef\thealgruleheight{\dimexpr \the\ht\statebox+1pt\relax}%
	\edef\thealgruledepth{\dimexpr \the\dp\statebox+1pt\relax}%
	\ifdim\thealgruleheight<.75\baselineskip
	\def\thealgruleheight{\dimexpr .75\baselineskip+1pt\relax}%
	\fi
	\ifdim\thealgruledepth<.25\baselineskip
	\def\thealgruledepth{\dimexpr .25\baselineskip+1pt\relax}%
	\fi
	\State #1%
	\def\thealgruleheight{\dimexpr .75\baselineskip+1pt\relax}%
	\def\thealgruledepth{\dimexpr .25\baselineskip+1pt\relax}%
}

\newtheorem{myDef}{Definition}

\begin{document}\sloppy

\def\x{{\mathbf x}}
\def\L{{\cal L}}

\title{Fast Registration for cross-source point clouds by using weak regional affinity and pixel-wise refinement}
%
 
\name{Xiaoshui Huang$^{1}$, Lixin Fan$^{2}$, Qiang Wu$^{1}$, Jian Zhang$^{1}$, Chun Yuan$^{3}$}
\address{1. GBDTC, University of Technology Sydney, Australia \\ 2. Nokia Technologies, Finland \\3. Graduate school at Shenzhen, Tsinghua University, China     }
\maketitle

\begin{abstract}
	Many types of 3D acquisition sensors are emerged in  recent  years and point cloud has been widely used in many areas. Accurate and fast registration of cross-source 3D point clouds from different sensors is an emerged research problem in computer vision. This problem is extremely challenging because cross-source point clouds contain mixture of various variances, such as density, partial overlap, large noise and outliers, viewpoint changing. In this paper, an algorithm is proposed to align cross-source point clouds with both high accuracy and high efficiency. There are two main contributions: firstly, two components, the weak region affinity and pixel-wise refinement, are proposed to maintain the global and local information of 3D point clouds. Then, these two components are integrated into an iterative tensor-based registration algorithm to solve the cross-source point cloud registration problem. We conduct  experiments on synthetic cross-source benchmark dataset and real cross-source datasets. Comparison with six state-of-the-art methods, the proposed method obtains both higher efficiency and accuracy.
\end{abstract}

\section{Introduction}
\label{introduction}
With  the  rapid  development  of  3D  data acquisition technologies, point clouds are becoming  an effective  way to express the surfaces of 3D objects and scenes. Because 3D point clouds have many advantages over 2D images, for example easily extracting spatial geometric, shape and pose information, point clouds become popular to be used in many fields and applications. Some current vision systems even contain more than one type of sensors, such as LiDAR and stereo camera in autonomous driving vision system. The existing applications show improved performance by using fused point clouds from different types of sensors. However, fusion of cross-source point clouds are very challenging because they contain mixture of various differences, such as different density, noise, outliers, viewpoint changing and missing data. The detail of the differences is explained in \cite{CSGM}\cite{huang2018coarse}. In this paper, a registration algorithm is proposed to deal with cross-source point cloud fusion problem by considering weak region affinity and pixel-wise refinement. 

\begin{myDef}
	\label{def1}
	Same-source point clouds are homogeneous point clouds from same types of sensors. Cross-source point clouds are heterogeneous point clouds from different types of sensors.
\end{myDef}

\begin{figure}[t]
	\centering
	\includegraphics[height=3cm,width=80mm]{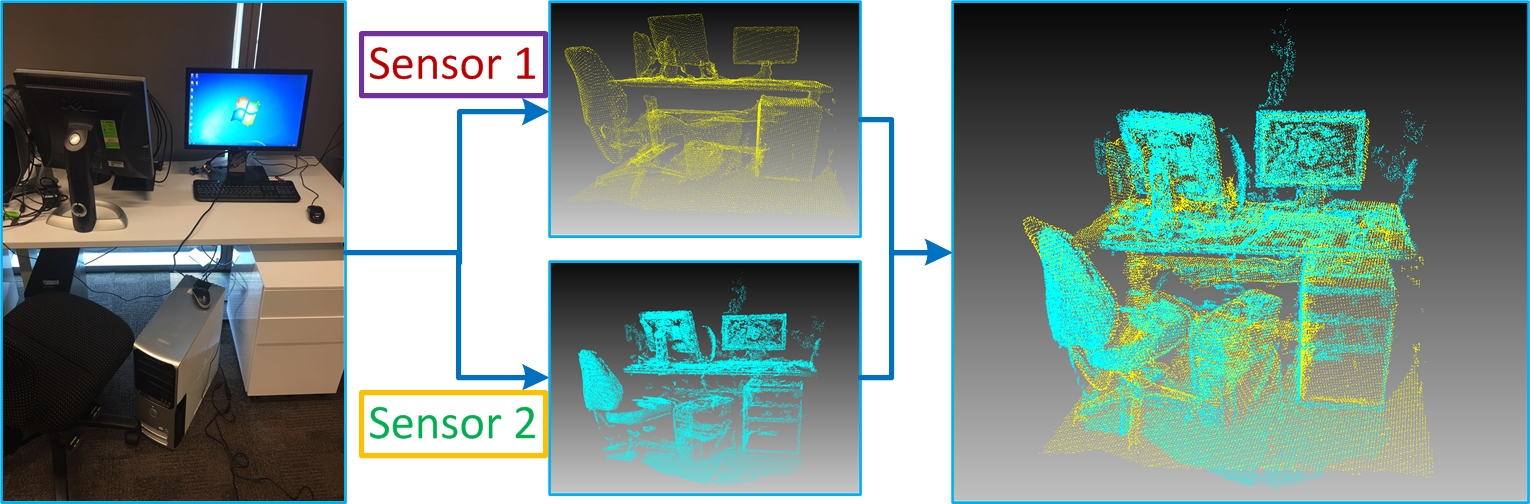}
	\caption{Two cross-source point clouds are captured about a same indoor scene. They contain mixture of variants of density, outliers, noise and missing data. The proposed algorithm can successfully align them with high accuracy and efficiency.}
	\label{start}
\end{figure}

There are several attempts to solve the cross-source point cloud registration. CSGM \cite{CSGM}  converts registration problem into graph matching problem. Then, ICP refinement (local) is done based on the graph matching results (global). \cite{peng2014street} proposes a coarse-to-fine (global-to-local) strategy to solve the cross-source point cloud. \cite{tvcg} utilizes RANSAC and ICP to global align and local refine the cross-source point cloud registration. However, all the existing methods either cannot solve the cross-source point cloud registration accurately or cannot solve the problem efficiently. All the existing registration methods assume the strong structure consistency between two point-cloud sets being registered. However, it is not the case of cross source point cloud registration in which structure consistence is weak. The reliable correspondences which can be located between two sets are sparse. Larger inconsistency caused by different point cloud density, noise model and various outliers significantly degrade the estimation on rigid transformation. 

According to the observation (One example is Figure \ref{start}), any two given cross-source point clouds still hold the intrinsic structure similarity although it is weak and degraded by the various noise, inconsistent point cloud density and outliers. The proposed method is able to discover the salient but weak geometric structure affinity (not statistic feature affinity) by joining sufficient number of local matching i.e. weak regional affinity. In order to adjust the mismatching in weak regional affinity search process, a pixel-wise refinement process is proposed. It is similar to global and local strategy in previous existing methods. Different to previous separated streams, in the proposed method, these two processes i.e. co-affinity search and pixel-wise refinement is unified together. Thus, it can be sorted out in an uniform optimization process. Compared to separated streams, the advantages of uniform optimization process is that interaction between local and global components will be given to the optimization process by jointly considering both local and global information in one unified process.

To mathematically formalize the above motivation, we need a tool that is a generalization of vectors and matrices. We select tensor as the mathematic tool to assemble this idea, because tensor provides an elegant and unified mathematic format to assemble the global weak regional affinity and local pixel-wise refinement. The weak regional affinity is triplet constraint that is stored in third-order tensors. The pixel-wise refinement is point-point or point-plane residual error that is stored in first-order tensor. Then, the correspondent finding problem can be optimized in a unified whole potential tensor space. Compared to previous separated global and local strategy, these components are interacted in the whole optimization process thanks to the tensor optimization. To solve the final registration problem, instead of doing tensor optimization once, we propose an iterative tensor optimization solution and a new energy function is formulated to obtain optimal geometric transformation. During the iterative optimization solution, in order to interact between transformation matrix estimation and correspondence estimation, the geometric transformation $T$ is integrated into tensor optimization and the tensor space will be updated when the $T$ is updated.  When the energy function obtains convergence, both the correspondence and the geometric transformation are optimized.


The contributions of this paper are three aspects: 
\begin{itemize}
	\item (1) two components, weak regional affinity and pixel-wise refinement, are proposed to keep global and local information in cross-source point clouds where structures are usually weak. We assemble them into tensors so that interaction between global and local information becomes possible. 
	\item (2) An registration algorithm is proposed to integrate these two components to solve the cross-source point cloud registration. We iteratively updated tensor space so that interaction between correspondence estimation and transformation estimation is considered. 
	\item (3) Comprehensive comparison experiments are conducted on cross-source point cloud registration problem.
\end{itemize}

\section{Proposed algorithm}
\label{method}
Inspired by \cite{chui2003new}, we recognize a point-point match as a correct match only if the selected triplet points are matched. We regard these triplet points matched constraint as weak region affinity constraint because the triplet point selection is an elegant way to remain the salient structure in the rigid+scale geometric transformation problem. We assemble weak region affinity into three-order tensor. Also, the point-point match is regarded as pixel-wise refinement and can be assembled as first-order tensor. In our method, the three-order tensor and the first-order tensor are integrated into a tensor optimization framework. Then, an efficient power iteration solution is proposed to solve the tensor optimization. In order to be more robust to geometric deformations, we simultaneously solve the optimal correspondence $X$ as well as the optimal transformation $T$.
Because the proposed method considers geometric constraint and tensor optimization, we define it geometric constraint tensor-based registration (GCTR).

\subsection{Definition of pixel-wise refinement and  weak region affinity}
\label{definition}
In the following, we will introduce how to formulate the above pixel-wise refinement and weak regional affinity into tensors. First-order tensor and third-order tensor are utilized to store pixel-wise refinement and weak regional affinity separately.  In the following, we suppose Point cloud $C_1$ has $N_1$ points and point cloud $C_2$ has $N_2$ points.

\begin{figure}[ht]
	\centering
	\includegraphics[height=8mm,width=80mm]{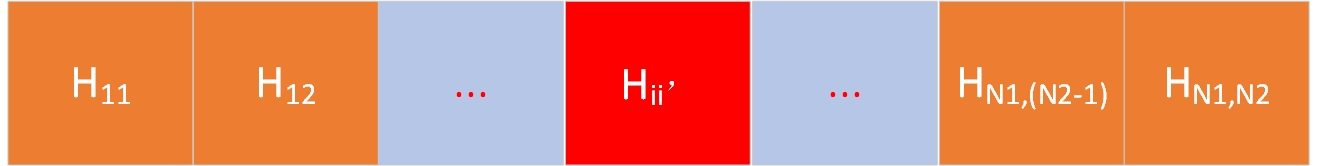}
	\caption{First-order tensor $\bar{H}_\alpha$ to represent point-to-point similarity in two point clouds with $N_1$ and $N_2$ points. The first-order tensor is constructed by concatenating the columns of $H$. To index a value of the first-order tensor $\bar{H}_\alpha$, for example the red box $H_{ii'}$, it is $\bar{H}_{(i + i'\times N_1)}$. }
	\label{tensor1}
\end{figure}

\begin{itemize}
	\item \textbf{Pixel-wise Refinement:} Pixel-wise refinement is the potential point-to-point correspondence. In our algorithm, first-order tensor is used to store pixel-wise refinement represents correspondent similarity. For correspondent similarity, it is computed as the Euclidean distance between pixel-wise point pair:
	\begin{equation}
	{H}_{ii'}=exp(-\Vert f_i-f_{i'} \Vert^2)
	\label{h1}
	\end{equation}
	where  $H$ is a $N_1 \times N_2$ correspondent point similarity matrix, $i \in [0,N_1)$ and $i' \in [0,N_2)$. Each element of $H$ stores the similarity of point $i$ of point cloud $C_1$ and point $i'$ of point cloud $C_2$. 
	
	For the first-order tensor $\bar{H_\alpha}$, it is a $N_1N_2$ vector by concatenating the rows of a similarity matrix $H$. The index conversion between $\bar{H}_\alpha$ to $H_{ii'}$ is $ i = i + i'\times N_1$ (See Figure \ref{tensor1}).  $f_i$ is the feature vector of point $P_i$ in point cloud $C_1$ and $f_{i'}$ is the feature vector of point $P_{i'}$ in point cloud $C_2$.  For feature vector, we use 3D point coordinate. The first-order tensor stores local information.
	
	\item \textbf{Weak Regional Affinity:} Weak regional affinity is the potential triplet-to-triplet correspondence(triplet points are a simple region). In our algorithm, third-order tensor is used to store weak regional affinity. In particular, triplet points are selected and are used to represent weak salient structure of cross-source point cloud (triplet points selection is detailed in Section \ref{implement}). To estimate the correspondence between triplet points, we need to compute the similarity of these triplets. The similarity is computed by 
	\begin{equation}
	\begin{aligned}
	{H}_{ii'jj'kk'}=exp(-\Vert f_{ijk}-f_{i'j'k'} \Vert^2)
	\label{h3}
	\end{aligned}
	\end{equation}
	where ${H}_{ii'jj'kk'}$ is a 6D supersymmetric tensor such as invariant under permutations of indices in $(i,j,k)$ or ${i',j',k'}$. Each element of the 6D tensor stores the similarity of the two triplets. Points ($P_i, P_j, P_k$) and ($P_{i'}, P_{j'}, P_{k'}$) are two triplet points with the correspondent relations based on their orders. In particular, the above order represent Point $P_i$ of triplet 1 is correspondent to point $P_{i'}$ of triplet 2, and the same to point correspondence of $(P_j, P_{j'})$ and $(P_k,P_{k'})$. $\bar{H}_{\alpha\beta\gamma}$ is a three-order tensor of size  $(N_1N_2)^3$ which is rewritten from tensor $H_{ii'jj'kk'}$. Because triplet points are selected as large triangles, the third-order tensor stores global information.
	
	\item \textbf{Triplet similarity: }For each triplet, we compute cosine value of three inner angles of the correspondent triangle combined by the triplet. Then, a descriptor ($f_{i_sj_sk_s}$ , $s\in\{1,2\}$) is formed to describe the triplet. The similarity between triplets are computed by using their descriptors $f_{i_sj_sk_s}$. Using this similarity computation strategy, all the elements of the above three-order tensor can be computed. In the three-order tensor, each dimension reflects the potential of point-point correspondence. Therefore, three dimensions are same and the node correspondent will be permutation of all points between  two point clouds. Therefore, the tensor is a symmetric tensor.
\end{itemize}

\begin{figure}[ht]
	\centering
	\includegraphics[height=30mm,width=80mm]{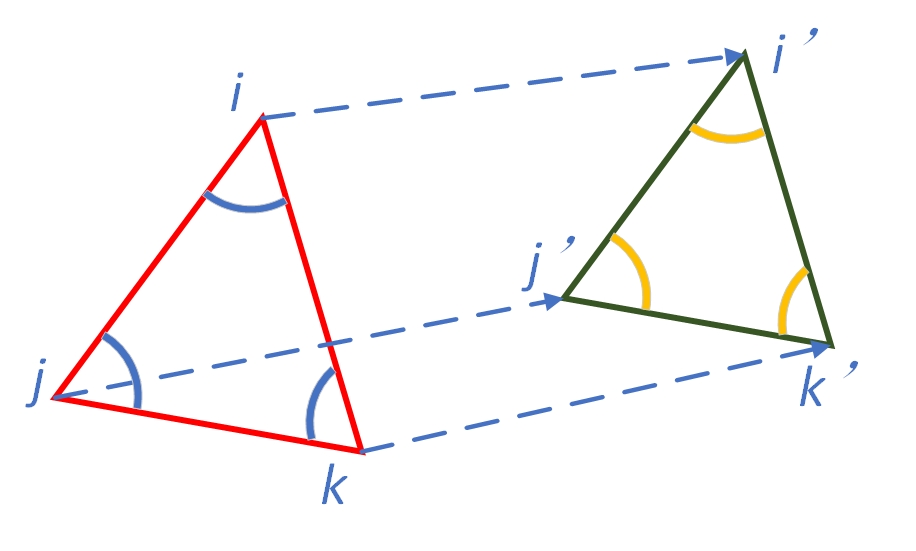}
	\caption{Similarity of each triplet point , which is computed by two descriptor of two triplets. For the descriptor of each triplet, it is three cosine value of three inner angles of the triangle.}
	\label{similarity}
\end{figure}

\subsection{Geometric Constraint Tensor-based registration method (GCTR)}

The goal of our method is to find the optimal transformation matrix between cross-source point clouds. One key step is to finding point-point matching by considering triplet constraint. This is equal to finding the best matching solution in the whole tensor space. In this section, we integrate the above two components into an unified framework and solving the registration problem by maximizing the following objective function:

\begin{eqnarray}
\label{scoreFour}
\begin{aligned}
S&=\sum\limits_{i,j,k}H_{ii'jj'kk'}(T)X_{ii'}X_{jj'}X_{kk'} + {H_{ii'}(T)X_{ii'}}\\ 
&=\bar{H}_{\alpha\beta\gamma}(T)\otimes_3\bar{X}\otimes_2\bar{X}\otimes_1\bar{X}+ \bar{H}_l(T)\otimes_1\bar{X}
\end{aligned}
\end{eqnarray}
where $X$ (correspondence matrix) and $T$ (transformation matrix) are two parameters need to estimate. $H_{ii'jj'kk'}$ is a 6D supersymmetric tensor. Each node pair's (i.e. $P_i^1$ and $P_{i'}^2$) similarity contributes a 2D dimension matrix to $H_{ii'jj'kk'}$, i.e. $P_i$ and $P_i'$ can form a 2D matrix, and $P_j$ and $P_j'$ can form another 2D matrix. $H_{ll'}$ is a 2D matrix to describe the pixel-wise similarity. $X$ is the $N_1\times N_2$ assignment matrix where $1$ means two points are matched and $0$ otherwise. $\bar{X}=vec(X)$ obtains $N_1N_2$ vector form of $X$  by concatenating the columns of $X$. $\bar{H_{\alpha\beta\gamma}}=vec(H_{ii'jj'kk'})$ is a three-order tensor of size $(N_1N_2)^3$ where each element represents the similarity of two triplets. It is a rewritten of tensor $H_{ii'jj'kk'}$. $\bar{H}_l=vec(H_{ii'})$ is vector form of $H_{ii'}$ by concatenating the columns of $H_{ii'}$, where each element represents the point-point similarity. With a geometric transformation $T(.)$ given, $\bar{H}_{ii'jj'kk'}(T)$ and $\bar{H}_{ii'}(T)$ are the two specific tensors and correspondence matrix $X$ can be estimated by tensor optimization. With an optimized $X$, transformation matrix $T$ can be estimated by singular-value decomposition (SVD).

For the scale computation, we compare triplet correspondent edges and compute the mean ratio as the final scale:

\begin{eqnarray}
\label{scale}
s=\frac{\sum\limits_{i=1}^{n-1}(r_{i}^a/r_{i}^b)}{n-1}
\end{eqnarray}
where $r_{i}^a$ is the length of point $A_i$ and point $A_{i+1}$ in point set $A$, $B_i$ is the correspondence of $A_i$ and $B_{i+1}$ is the correspondence of $A_{i+1}$, $r_{i}^b$ is the length of $B_i$ and point $B_{i+1}$. $n$ is the number of correspondent pairs, $a, b$ identifies the length in point set A and point set B.

\subsection{Power iteration solution}
The power iteration solution aims to optimizing the objective function \ref{scoreFour} by two iterative processes: the correspondence $X$ is optimized by firming the geometric transformation $T$ and the geometric transformation $T$ is optimized by firming the new correspondence $X$. Algorithm \ref{GCTR} shows the whole process.
\subsubsection{Optimization for the correspondence}
With specific geometric transformation $T$ given, the optimization of formulation \ref{scoreFour} is a tensor optimization problem. According to \cite{shi2013multi}, two terms in objective function \ref{scoreFour} are rank-1 tensors, so that the optimization can be formulated as $Rank-1 \ tensor \ approximation$ (R1TA) problem. Inspired by \cite{shi2013multi,duchenne2011tensor}, we use tensor power iteration to solve the above R1TA problems. Line 6-10 in Algorithm \ref{GCTR} shows the correspondence optimization procedures. 


\subsubsection{Optimization for the geometric transformation}
With the correspondences $X$ are given, the estimation of the geometric transformation $T$ is similar to ICP which can be solved in close-form solution. Suppose $A$ and $B$ is $n$ matched pairs from correspondence matrix $X$, $A$ is the points from point set $C_1$ and $B$ is the points from point set $C_2$. If $U_A$ is the mean point of $A$ and $U_B$ is the mean point of $B$, $UDV=svd(AB^T)$, the geometrical transformation $T$ is computed by  $R=UDV^T, t=U_A-s*R*U_B, s=({\sum\limits_{i=1}^{n-1}(r_{ai}/r_{bi})})/{n-1}$. In this step, the key elements are scale estimation and tensor update. For scale estimation computation, we use formulation \ref{scale} defined before. For tensor update, we use formulation of Line 9 in Algorithm \ref{GCTR}.

\begin{algorithm}
	\caption{Power iteration algorithm for cross-source point cloud registration. }
	\label{GCTR}
	\begin{algorithmic}[1]
		\Require Cross-source  point clouds.
		\Ensure Transformation matrices R,t,s
		\myState Salient structures extraction
		\Repeat
		\myState Wide baseline triplets selection in $C_1,C_2$.
		\myState Build different-order tensors.
		\Repeat
		\myState{$\beta^{(m)}= \sum\limits_{i,j,k}H_{3}\prod_{r}^{{i,j,k}}X_r^{(m)}$ }
		\myState{$\gamma^{(m)}= H_1X^{(m)}$}
		\myState{$S^{(m+1)}=\beta^{(m)}+\gamma^{(m)}$}
		\myState{$X^{(m+1)}=H_3\otimes X^{(m)}\otimes X^{(m)}+H_1$} 
		\myState{$X^{(m+1)} \leftarrow \frac{1}{\Arrowvert X\Arrowvert_2}X^{(m+1)}$}
		\Until convergence;	
		\myState{$A_1,B_1 \leftarrow \Arrowvert X^{(m+1)}\Arrowvert_1^r$}
		\myState{$A,B \leftarrow RANSAC(A_1,B_1)$}
		\myState{$R=UDV^T,t=U_A-s*R*U_B$}
		\myState{$s=\frac{\sum\limits_{i=1}^{n-1}(r_{ai}/r_{bi})}{n-1}$}
		\myState Update point cloud $C_2=(sRC_{2}+t)$ 	
		\Until convergence;			
	\end{algorithmic}
\end{algorithm}

\section{Implementation details}
\label{implement}
Initially, we introduce line 1 of Algorithm \ref{GCTR}. Following \cite{CSGM}, we use supervoxel segmentation method \cite{papon2013voxel} to segment the point clouds and use the central points of these segments as the salient structures. 

\textbf{Triplet point selections: } Inspired by \cite{super4pcs}, we randomly select triangles satisfying wide baseline strategy and use the three nodes of these triangles as triplet points. 

Then, we introduce line 3-4 of Algorithm \ref{GCTR}.  Inspired by \cite{super4pcs}, we select triplets satisfying wide baseline strategy, so that they are more likely to be global aligned.  In this algorithm, we use wide baseline strategy to randomly select $N_1N_2$ large triangles in point cloud 1. We define the large triangles as three edges of the triangle are large than 50\% of the overlapping 3D containing voxel's the diameter. That guarantees the selected triangles are large triangles and make the final registration more prone to globally registered. For the overlapping ratio, if there is unknown, we automatically search as ratio is 0.25, 0.5, 0.75, 1.0.

\section{Experimental results}
\label{experiments}
In this section, we conduct thoroughly comparison experiments on synthetic and real datasets. Firstly, we compare on a synthetic cross-source benchmark datasets. Several state-of-the-art registration methods have been run on it and compared with the proposed method. Secondly, we compare the performance on real cross-source point clouds.


\subsection{Experiments setup}
The proposed algorithm is implemented by using standard C and Matlab. All the comparison experiments are executed on an I5 CPU, 8GB memory computer. We select ICP \cite{icp}, GO-ICP \cite{goicp}, Super-4PCS\cite{super4pcs}, CPD\cite{cpd}, JR-MPC\cite{jrmpc} and CSGM\cite{CSGM} as the comparison methods. Most of the existed state-of-the-art registration methods are focus on same-source data and designed to solve SE(3) transformation. They have not designed for scale variation. To compare fairly, we conduct an automatically scale normalization for all the other methods by following \cite{huang2016coarse}, which is assuming the size of the point cloud's 3D containing voxel is the same. Because JR-MPC becomes not practical when the point number increases significantly, we uniformly down-sample the point cloud to approximately 2000 points for JR-MPC. 

For the evaluation, we compute the Frobenius Norm of the transformation matrix difference (TM) between ground-truth and estimation. The lower the value is, the higher accuracy the method achieves.

\begin{figure}[htb]
	\centering
	\includegraphics[width=\linewidth]{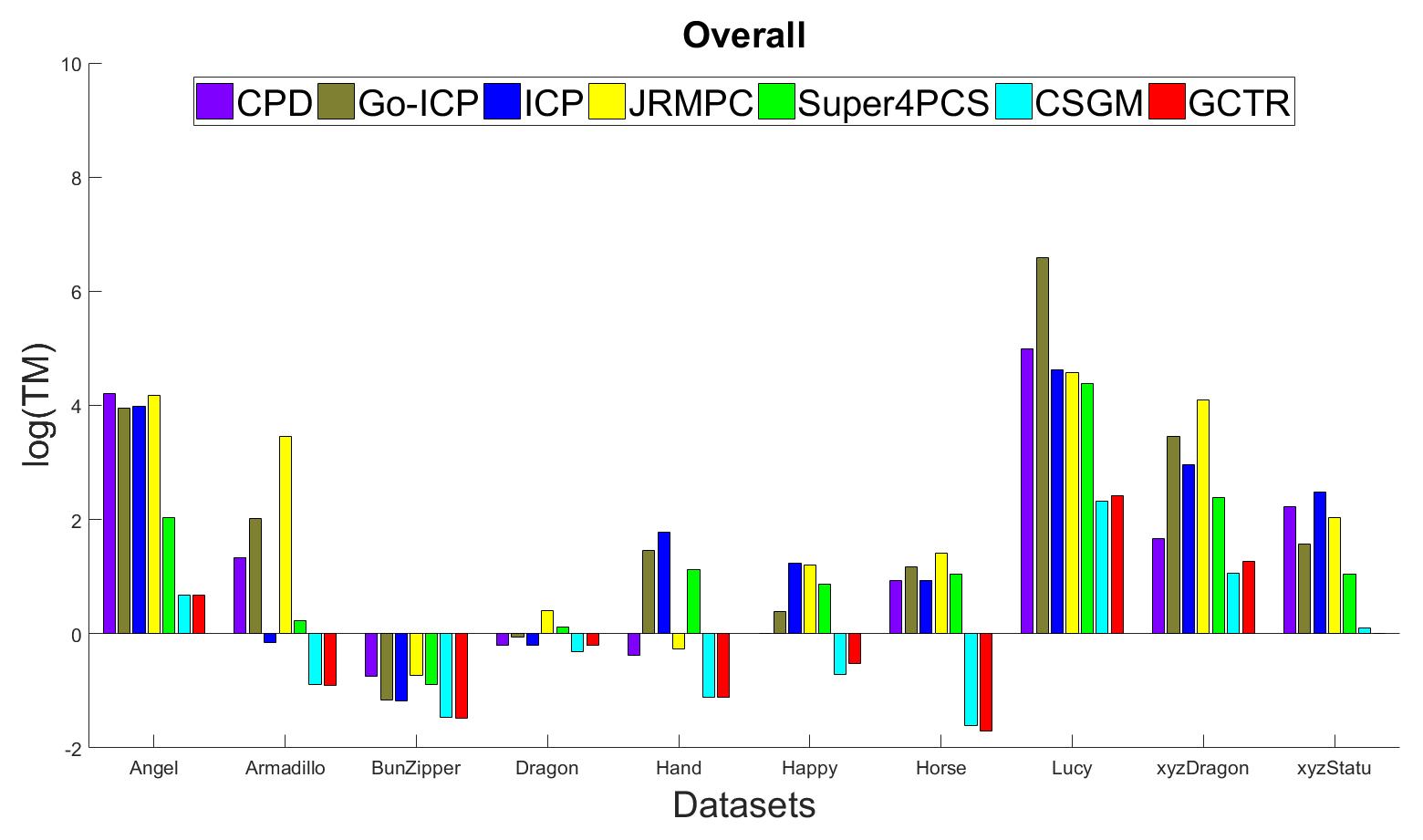}
	\caption{Overall performance (rotation,transformation and scale) on synthetic cross-source benchmark dataset.}
	
	\label{syndata_overall}
\end{figure}

\subsection{Synthetic cross-source benchmark dataset}
This section will demonstrate the ability of the proposed method on the cross-source benchmark dataset. For the construction of benchmark dataset, following \cite{CSGM}, we start from Stanford 3D Scanning Models  \footnote{http://graphics.stanford.edu/data/3Dscanrep/} and simulate ten sets of cross-source benchmark dataset. Each set of cross-source dataset contains source A and source B which simulate cross-source problems (discussed in Section \ref{introduction}).  
\begin{table}[h]
	\begin{center}
		\begin{tabular}{p{1.5cm}|p{0.5cm}p{0.5cm}p{0.5cm}|p{1.2cm}|p{1.0cm}}
			\hline
			Method & R & T & S& log(TM)& time(s)\\
			\hline\hline
			CPD      & 2.82 & 64.4 & 0.24 & 4.21 & 435\\
			Go-ICP   & 2.7  & 48.8 & 0.06 & 3.94 & 73.8\\
			ICP      & 3.38 & 50.1 & 0.06 & 3.98 & 103\\
			JR-MPC   & 1.84 & 63.3 & 0.06 & 4.18 & 255\\
			Super4PCS& 0.22 & 3.41 & 0.06 & 1.3  & 333\\
			CSGM & \textit{0.064} & \textit{1.85} & 0.06 & \textit{0.68}  & \textit{3024}\\
			GCTR     & \textit{0.069} & \textit{1.89} & \textbf{0.03} &\textit{0.69}&\textbf{61.6}\\
			\hline
		\end{tabular}
	\end{center}
	\caption{Comparison on Angle benchmark dataset.}
	\label{tableAngle}
\end{table}

\begin{table*}[ht]
	\begin{center}
		\centering
		\begin{tabular}{p{1.5cm}|p{0.8cm}p{1.4cm}p{0.8cm}p{1.4cm}p{1.6cm}p{0.8cm}p{0.8cm}p{0.8cm}p{1.0cm}}
			\hline
			Method & CPD & Go-ICP & ICP& JR-MPC& Super4PCS & CSGM & GCTR\\
			\hline\hline
			Runtime(s)& 650 & 363& 1491 &324 &1941 & 4648  &\textbf{287}\\
			\hline
		\end{tabular}
	\end{center}
	\caption{Average running time on the 10 pairs of cross-source benchmark datasets.}
	\label{averagetime}
\end{table*}


We select \textit{Angle} as examples for detailed quantitative evaluation on the cross-source benchmark datasets. We evaluate the \textit{Translation(T), Rotation(R), Scale (S)} error separately, and compare the error of  \textit{RMSE (log(TM))}. Also, we compare the runtime on this dataset. Table \ref{tableAngle} shows the evaluation results. It shows the GCTR obtains comparable accuracy to CSGM while shows much faster than CSGM. Compared to other methods, the proposed method obtains higher performance on accuracy and efficiency. 

Figure \ref{syndata_overall} shows the comparison results on whole datasets. We can see that our method obtains highest accuracy and robust results in all datasets. CSGM obtains comparable accuracy to the proposed method. In the other comparison methods, Super4PCS achieves the second performance for most datasets. An interesting phenomenon is that ICP obtains second accuracy than all other comparison methods in dataset Horse. Dataset horse is part of a horse and points arrange on a smooth surface. Our interpretation is that ICP shows some ability in aligning the cross-source point clouds when there is no scale variation and the initialization is very well and no large disorganized outliers. That is the reason why ICP is used as the final refinement step in many applications on same-source registration.


To compare the efficiency, we compute the average runtime on the 10 cross-source datasets. Table \ref{averagetime} show the proposed method is much faster than other compared methods. Although CSGM obtains comparable accuracy to the proposed method, our method achieves much higher efficiency than CSGM (about 16 times faster).

\begin{figure}[ht]
	\centering
	\includegraphics[height=5cm,width=\linewidth]{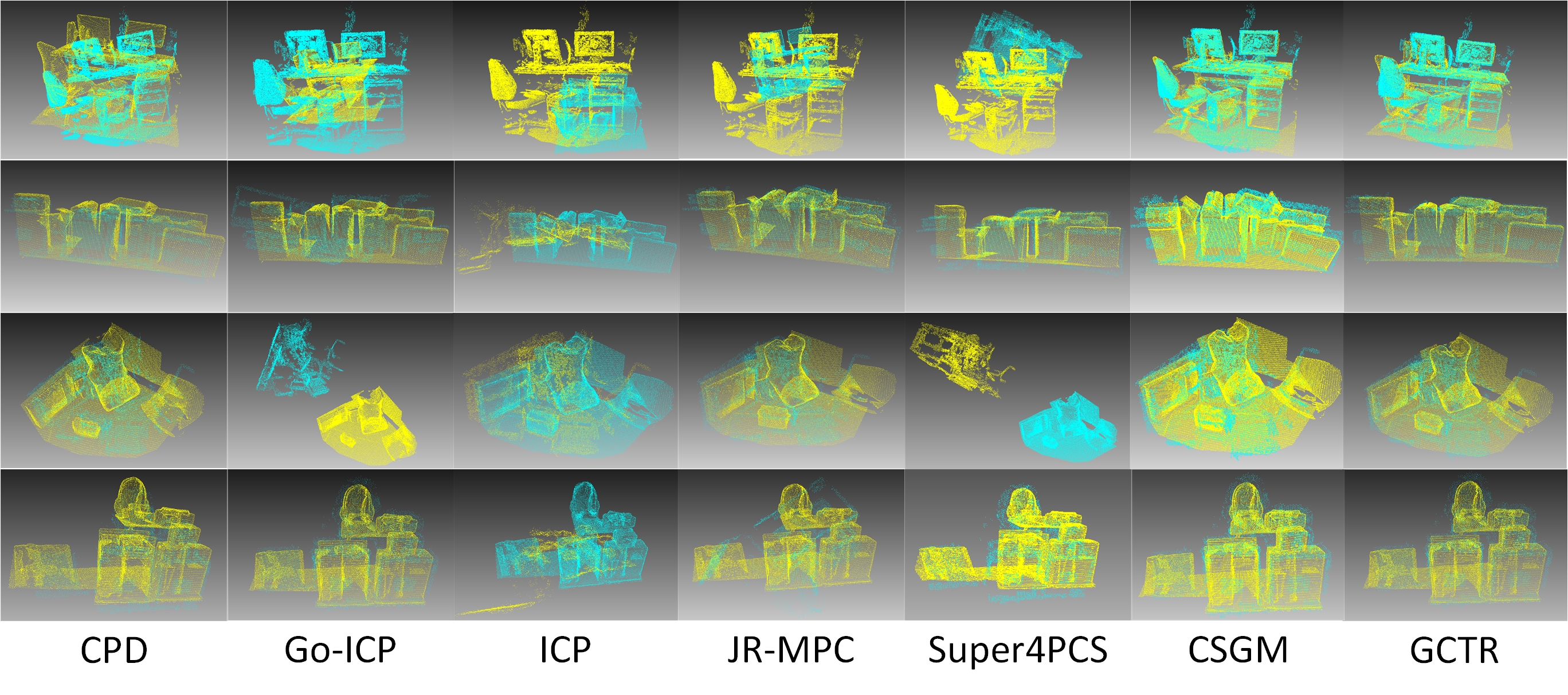}
	\caption{Visual registration results of real cross-source point clouds.}
	\label{real}
\end{figure}
\subsection{Real cross-source point clouds}
We capture the cross-source point cloud dataset by using KinectFusion and VSFM . KinectFusion outputs point cloud directly while VSFM reconstructs 3D point cloud from the images taking by mobile phone camera. We captured more than 30 datasets for different indoor scenes and the proposed methods obtain promising results in all the datasets.

Figure \ref{real} shows the selected datasets and compares with other methods. The results illustrate our method obtain robust and visually correct registration results in real application.  For all these datasets, the proposed methods align at a fast speed in approximately 120 seconds which is much faster than other methods. Although sometimes JR-MPC and Go-ICP obtains registration results similar to our methods, their computation and memory complexity are very high. Our method can align the real cross-source point cloud accurately at a fast speed.

\section{Conclusion}
In this paper, we propose a fast registration method for cross-source point clouds registration. We have done two main works: firstly, weak regional affinity and pixel-wise refinement are proposed to keep global and local information in cross-source point clouds where structures are usually weak; secondly, an unified algorithm is proposed to integrate these two components to solve the cross-source point cloud registration. The experimental results show that the proposed method aligns the challenging cross-source point cloud fast and accurately.

\bibliographystyle{IEEEbib}
\bibliography{IEEEfull}

\begin{thebibliography}{10}

\bibitem{CSGM}
X.~Huang, J.~Zhang, L.~Fan, Q.~Wu, and C.~Yuan,
\newblock ``A systematic approach for cross-source point cloud registration by
  preserving macro and micro structures,''
\newblock {\em IEEE Transactions on Image Processing}, vol. 26, no. 7, pp.
  3261--3276, July 2017.

\bibitem{huang2018coarse}
X.~Huang, J.~Zhang, Q.~Wu, L.~Fan, and C.~Yuan,
\newblock ``A coarse-to-fine algorithm for matching and registration in 3d
  cross-source point clouds,''
\newblock in {\em IEEE Transactions on Circuits and Systems for Video
  Technology}. IEEE, 2018, vol.~28, pp. 2965--2977.

\bibitem{peng2014street}
Furong Peng, Qiang Wu, Lixin Fan, Jian Zhang, Yu~You, Jianfeng Lu, and Jing-Yu
  Yang,
\newblock ``Street view cross-sourced point cloud matching and registration,''
\newblock in {\em Image Processing (ICIP), 2014 IEEE International Conference
  on}. IEEE, 2014, pp. 2026--2030.

\bibitem{tvcg}
N.~Mellado, M.~Dellepiane, and R.~Scopigno,
\newblock ``Relative scale estimation and 3d registration of multi-modal
  geometry using growing least squares,''
\newblock {\em IEEE Transactions on Visualization and Computer Graphics}, vol.
  22, no. 9, pp. 2160--2173, Sept 2016.

\bibitem{chui2003new}
Haili Chui and Anand Rangarajan,
\newblock ``A new point matching algorithm for non-rigid registration,''
\newblock {\em Computer Vision and Image Understanding}, vol. 89, no. 2, pp.
  114--141, 2003.

\bibitem{shi2013multi}
Xinchu Shi, Haibin Ling, Junling Xing, and Weiming Hu,
\newblock ``Multi-target tracking by rank-1 tensor approximation,''
\newblock in {\em Proceedings of the IEEE Conference on Computer Vision and
  Pattern Recognition}, 2013, pp. 2387--2394.

\bibitem{duchenne2011tensor}
Olivier Duchenne, Francis Bach, In-So Kweon, and Jean Ponce,
\newblock ``A tensor-based algorithm for high-order graph matching,''
\newblock {\em IEEE transactions on pattern analysis and machine intelligence},
  vol. 33, no. 12, pp. 2383--2395, 2011.

\bibitem{papon2013voxel}
Jeremie Papon, Alexey Abramov, Markus Schoeler, and Florentin Worgotter,
\newblock ``Voxel cloud connectivity segmentation-supervoxels for point
  clouds,''
\newblock in {\em Proceedings of the IEEE conference on computer vision and
  pattern recognition}, 2013, pp. 2027--2034.

\bibitem{super4pcs}
Nicolas Mellado, Dror Aiger, and Niloy~J Mitra,
\newblock ``Super 4pcs fast global pointcloud registration via smart
  indexing,''
\newblock in {\em Computer Graphics Forum}. Wiley Online Library, 2014,
  vol.~33, pp. 205--215.

\bibitem{icp}
Paul~J Best and Neil~D McKay,
\newblock ``A method for registration of 3-d shapes,''
\newblock {\em IEEE Transactions on pattern analysis and machine intelligence},
  vol. 14, no. 2, pp. 239--256, 1992.

\bibitem{goicp}
J.~Yang, H.~Li, D.~Campbell, and Y.~Jia,
\newblock ``Go-icp: A globally optimal solution to 3d icp point-set
  registration,''
\newblock {\em IEEE Transactions on Pattern Analysis and Machine Intelligence},
  vol. PP, no. 99, pp. 1--1, 2016.

\bibitem{cpd}
Andriy Myronenko and Xubo Song,
\newblock ``Point set registration: Coherent point drift,''
\newblock {\em IEEE transactions on pattern analysis and machine intelligence},
  vol. 32, no. 12, pp. 2262--2275, 2010.

\bibitem{jrmpc}
Georgios~Dimitrios Evangelidis and Radu Horaud,
\newblock ``Joint alignment of multiple point sets with batch and incremental
  expectation-maximization,''
\newblock {\em IEEE transactions on pattern analysis and machine intelligence},
  vol. 40, no. 6, pp. 1397--1410, 2018.

\bibitem{huang2016coarse}
Xiaoshui Huang, Jian Zhang, Qiang Wu, Lixin Fan, and Chun Yuan,
\newblock ``A coarse-to-fine algorithm for registration in 3d street-view
  cross-source point clouds,''
\newblock in {\em 2016 International Conference on Digital Image Computing:
  Techniques and Applications (DICTA)}. IEEE, 2016, pp. 1--6.

\end{thebibliography}

\end{document}